%% file: eccv2022submission.tex
\documentclass[runningheads]{llncs}
\usepackage{graphicx}

\usepackage{tikz}
\usepackage{comment}
\usepackage{amsmath,amssymb} 
\usepackage{color}
\usepackage{booktabs}
\usepackage[accsupp]{axessibility}  
\usepackage{indentfirst}
\usepackage{dsfont}
\usepackage{subfloat}
\usepackage{multirow}
\usepackage[pagebackref,breaklinks,colorlinks]{hyperref}

\newcommand\blfootnote[1]{%
  \begingroup
  \renewcommand\thefootnote{}\footnote{#1}%
  \addtocounter{footnote}{-1}%
  \endgroup
}

\usepackage[width=122mm,left=12mm,paperwidth=146mm,height=193mm,top=12mm,paperheight=217mm]{geometry}

\begin{document}
\pagestyle{headings}
\mainmatter
\def\ECCVSubNumber{125}  

\title{SeqFormer: Sequential Transformer for Video Instance Segmentation} 

\titlerunning{ECCV-22 submission ID \ECCVSubNumber} 
\authorrunning{ECCV-22 submission ID \ECCVSubNumber} 
\author{Anonymous ECCV submission}
\institute{Paper ID \ECCVSubNumber}

\titlerunning{SeqFormer for Video Instance Segmentation}
%
\author{Junfeng Wu\inst{1*} \and
Yi Jiang\inst{2} \and
Song Bai\inst{2} \and
Wenqing Zhang\inst{1*} \and 
Xiang Bai\inst{1\dag}
}
\authorrunning{J. Wu et al.}
%
\institute{
$^{1}$ Huazhong University of Science and Technology
$^{2}$ Bytedance 
}

\maketitle

\blfootnote{
\noindent $*$ Work done during an internship at ByteDance. \\
$\dag$ Corresponding author: \href{mailto:xbai@hust.edu.cn}{\color{black}{xbai@hust.edu.cn}}
}

\begin{abstract}
In this work, we present SeqFormer for video instance segmentation. SeqFormer follows the principle of vision transformer that models instance relationships among video frames. Nevertheless, we observe that a stand-alone instance query suffices for capturing a time sequence of instances in a video, but attention mechanisms shall be done with each frame independently. To achieve this, SeqFormer locates an instance in each frame and aggregates temporal information to learn a powerful representation of a video-level instance, which is used to predict the mask sequences on each frame dynamically. Instance tracking is achieved naturally without tracking branches or post-processing.
On YouTube-VIS, SeqFormer achieves 47.4 AP with a ResNet-50 backbone and 49.0 AP with a ResNet-101 backbone without bells and whistles. Such achievement significantly exceeds the previous state-of-the-art performance by 4.6 and 4.4, respectively. In addition, integrated with the recently-proposed Swin transformer, SeqFormer achieves a much higher AP of 59.3. We hope SeqFormer could be a strong baseline that fosters future research in video instance segmentation, and in the meantime, advances this field with a more robust, accurate, neat model. The code is available at \href{https://github.com/wjf5203/SeqFormer}{https://github.com/wjf5203/SeqFormer}.

\end{abstract}

\section{Introduction}

\begin{figure}[tb]
\centering
\includegraphics[width = 0.6\textwidth]{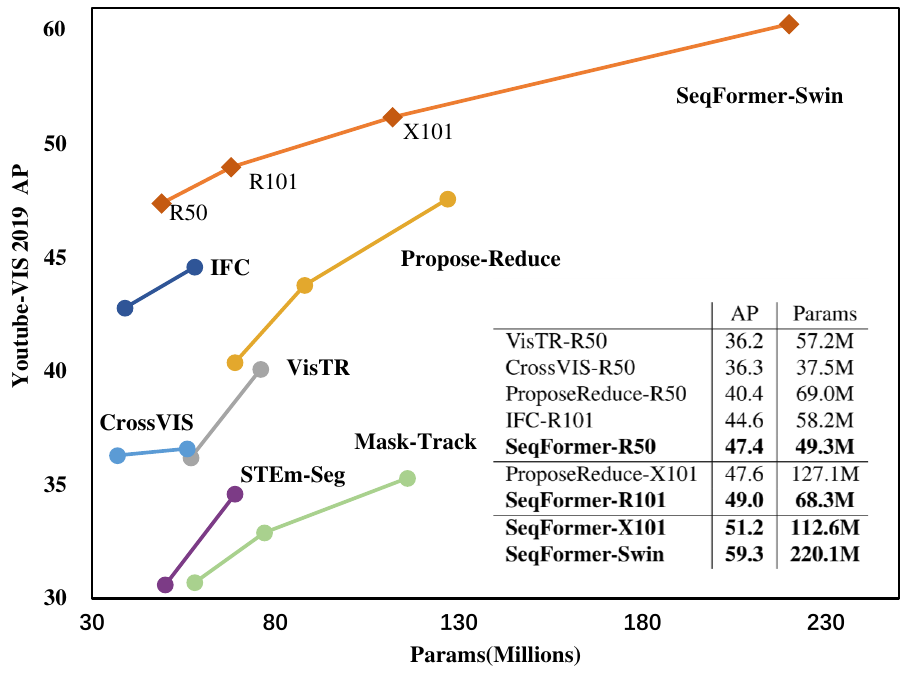}
\caption{\textbf{Performance \emph{vs.} Model Size.} All results are reported with single model and single-scale inference. SeqFormer significantly outperforms the previous method with similar parameters.}

\label{fig1}
\end{figure}

Video Instance Segmentation (VIS)~\cite{yang2019video,ovis} is an emerging vision task that aims to simultaneously perform detection, classification, segmentation, and tracking of object instances in videos. Compared to image instance segmentation~\cite{MaskRCNN}, video instance segmentation is much more challenging since it requires accurate tracking of objects across an entire video.

Previous VIS algorithms can be roughly divided into two categories.
The first mainstream follows the tracking-by-detection paradigm, extending image instance segmentation models with a tracking branch~\cite{yang2019video,cao2020sipmask,yang2021crossover,liu2021sg,1stYTVIS21,MSN}. These methods first predict candidate detection and segmentation frame-by-frame, and then associate them by classification~\cite{yang2019video,yang2021crossover} or re-identification~\cite{cao2020sipmask,liu2021sg} to track the instance through a video. Nevertheless, the tracking process is sensitive to occlusions and motion blur that are common in videos.
Another mainstream is to predict clip-level instance masks by taking a video clip~\cite{athar2020stem,MaskProp} or the entire video~\cite{IFC,VisTR} as input. It divides a video into multiple overlapping clips and generates mask sequences with clip-by-clip matching on overlapping frames. 
More recently, VisTR~\cite{VisTR} first adapts transformer~\cite{transformers} to VIS and uses instance queries to obtain instance sequence from video clips. After that, IFC~\cite{IFC} improves the performance and efficiency of VisTR by building communications between frames in a transformer encoder.

In this paper, we present Sequential Transformer (SeqFormer), which follows the principle of vision transformer~\cite{detr,VisTR} and models instance relationships among video frames. As in ~\cite{IFC}, we observe that a stand-alone instance query suffices although an object may be of different positions, sizes, shapes, and various appearances. Nevertheless, it is witnessed that the attention process shall be done with each frame independently, so that the model will attend to locations following with the movement of instance through the video. This observation aligns with the conclusion drawn in action recognition~\cite{zhao2018trajectory,MotionFormer}, where the 1D time domain and 2D space domain have different characteristics and should be handled in a different fashion.

Considering the movement of an instance in a video, a model is supposed to attend to different spatial locations following the motion of the instance.
We decompose the shared instance query into frame-level box queries for the attention mechanism to guarantee that the attention focuses on the same instance on each frame.
The box queries are kept on each frame and used to predict the bounding box sequences. Then the features within the bounding boxes are aggregated to refine the box queries on the current frame.
By repeating this refinement through decoder layers, SeqFormer locates the instance in each frame in a coarse-to-fine manner, in a similar way to Deformable DETR~\cite{deformableDETR}.

However, to mitigate redundant information from non-instance frames, those box queries are aggregated in a weighted manner, where the weights are end to end learned upon the box embeddings. The generated representation, which retains richer object cues, is used to predict the category and generate dynamic convolution weights of mask head.
Since the box sequences are predicted and refined in the decoder, SeqFormer naturally and succinctly establishes the association of instances across frames.

In summary, SeqFormer enjoys the following advantages:
\begin{itemize}
\item SeqFormer is a neat and efficient end-to-end framework. Given an arbitrary long video as input, SeqFormer predicts the classification results, box sequences, and mask sequences in one step without the need for additional tracking branches or hand-craft post-processing.

\item As shown in Fig.~\ref{fig1}, SeqFormer sets the new state-of-the-art performance on YouTube-VIS 2019 benchmark~\cite{yang2019video}. SeqFormer achieves 47.4 AP with a ResNet-50 backbone and 49.0 AP with a ResNet-101 backbone without bells and whistles. Such achievement significantly exceeds the previous state-of-the-art performance by 4.6 and 4.4, respectively. With a ResNext-101 backbone, SeqFormer achieves 51.2 AP, which is the first time that an algorithm achieves an AP above 50. In addition, integrated with the recently-proposed Swin transformer, SeqFormer achieves a much higher AP of 59.3.  
\item With the query decomposition mechanism, SeqFormer attends to locations following with the movement of instance through the video and learns a powerful representation for instance sequences.
\item The code and the pre-trained models are publicly available. We hope the SeqFormer, with the idea of making attention follow with the movement of object, could be a strong baseline that fosters future research in video instance segmentation, and in the meantime, advances this field with a more robust, accurate, neat model. 
\end{itemize}

\section{Related Work}
\label{sec:relatedwork}

\noindent\textbf{Image Instance Segmentation}
Instance Segmentation is the most fundamental and challenging task in computer vision, which aims to detect every instance and segment every pixel respectively in static images. 
Instance segmentation was dominated by Mask R-CNN architecture~\cite{MaskRCNN,PANet,maskscore} for a long time, Mask R-CNN~\cite{MaskRCNN} directly introduces fully convolutional mask head to Faster R-CNN~\cite{fastercnn} in a multi-task learning manner. Recently, one stage models~\cite{tensormask,solo,polarmask,tian2020conditional} emerged as excellent frameworks for instance segmentation. Solo~\cite{solo} and CondInst~\cite{tian2020conditional} propose one stage instance segmentation pipeline and achieve comparable performance. CondInst~\cite{tian2020conditional} proposes to dynamically generate the mask head parameters for each instance, which is used to predict the mask of the corresponding instance. QueryInst~\cite{queryinst} proposes a query based instance segmentation framework based on Sparse R-CNN~\cite{sparsercnn}, which also take advantage of the Dynamic mask head. 
Dynamic mask head can be efficiently adopted into video segmentation tasks because 
instances with the same identity on different frames can share the same mask head parameters.

\noindent\textbf{Video Instance Segmentation.}
Video instance segmentation is extended from the traditional image instance segmentation, and aims to simultaneously segment and track all object instances in the video sequence.
The baseline method MaskTrack R-CNN~\cite{yang2019video} is built upon Mask R-CNN~\cite{MaskRCNN} and introduces a tracking head to associate each instance in the video.
SipMask~\cite{cao2020sipmask} proposes a spatial preservation module to generate spatial coefficients for mask predictions based on the one-stage FCOS~\cite{FCOS}.
STMask~\cite{STMask} proposes a spatial feature calibration to extract features for frame-level instance segmentation on each frame, and further introduces a temporal fusion module to aggregate temporal information from adjacent frames.
STEm-Seg~\cite{athar2020stem} models a video clip as a single 3D spatial-temporal volume and enables inference procedure based on clustering.
CrossVIS~\cite{yang2021crossover} proposes a learning scheme that uses the instance feature in the current frame to pixel-wisely localize the same instance in other frames.
MaskProp~\cite{MaskProp} and Propose-Reduce~\cite{ProposeReduce} take advantage of mask propagation, which can achieve high performance, but it is very computationally intensive.


\noindent\textbf{Transformers.}
Transformer~\cite{transformers} was first proposed for the sequence-to-sequence machine translation task and became the basic component in most Natural Language Processing tasks. Recently, Transformer has been successfully applied in many visual tasks such as Object detection~\cite{detr,deformableDETR,sparsercnn}, segmentation~\cite{segformer,SETR,VisTR,referformer}, tracking~\cite{trackformer,transtrack,STARK}, video recognition~\cite{MotionFormer,uniformer,arnab2021vivit,ActionDetection}.
VIT\cite{VIT} firstly applies transformer in image recognition and model an image as sequence of patches, which achieves comparable performance with traditional CNN architecture. 
DETR~\cite{detr} proposes a new detection paradigm upon transformers, which simplifies the traditional detection framework and abandons the hand-crafted post-processing module. Deformable DETR~\cite{deformableDETR} achieves better performance by using local attention and multi-scale feature maps. VisTR~\cite{VisTR} is the first method that adapts Transformer to the VIS task. However, VisTR has a fixed number of input queries hardcoded by video length and maximum number of instances. Each query corresponds to an object on every single frame. In our method, instances with the same identity share a same query, which aggregates information across the video and learn a global feature representation for efficient segmentation. 
IFC~\cite{IFC} improves the performance and efficiency of VisTR by building communications between frames in the transformer encoder instead of flatting the space-time features into one dimension, but it still flatten the space-time features for the transformer decoder. Our model is designed to carry out the instance feature capturing independently on different frames, which makes the model attend to locations following with the movement of instance through the video.

\begin{figure*}[tb]
\centering
\includegraphics[width=1\linewidth]{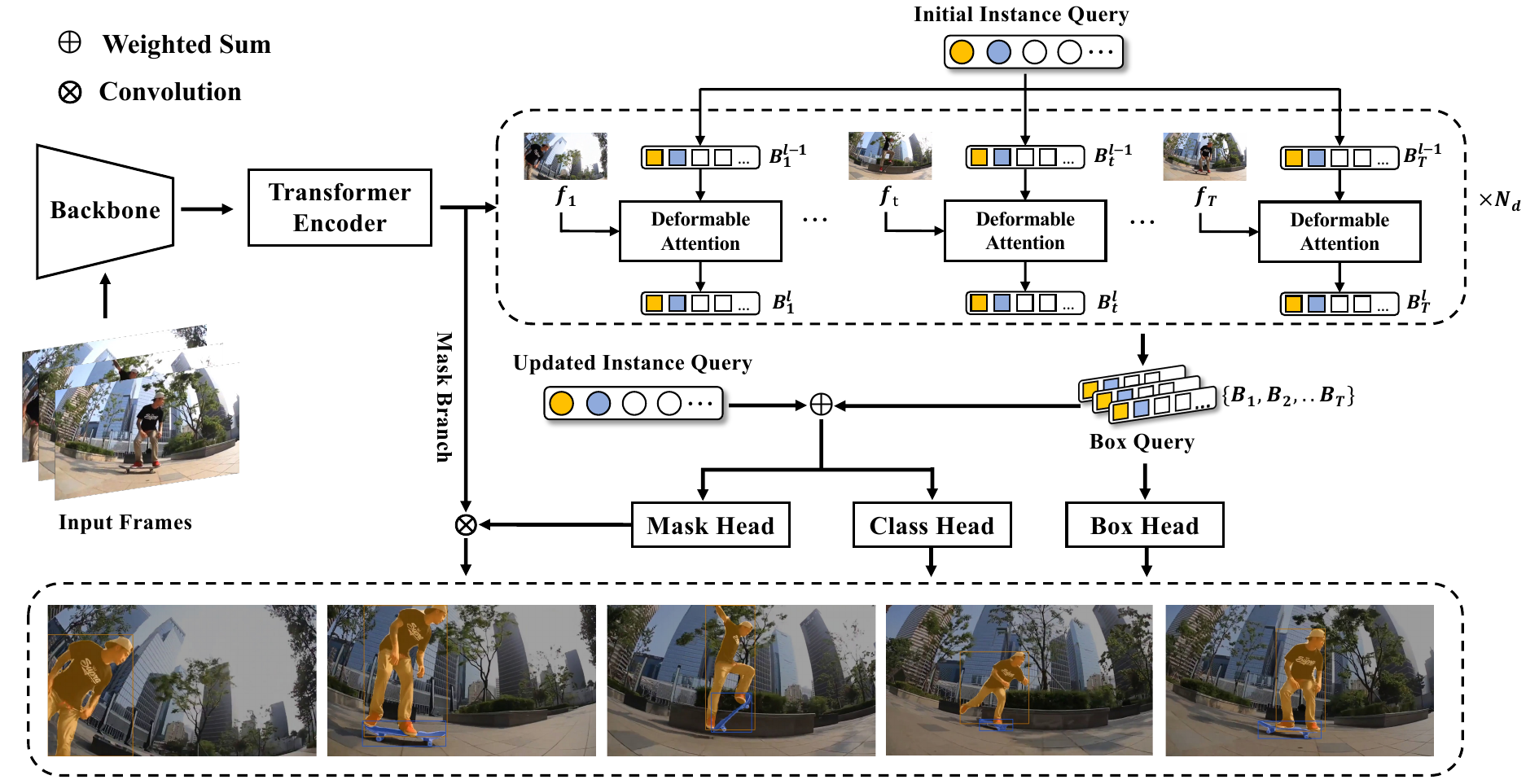}
\caption{The overall architecture of SeqFormer. Given the feature maps of input frames, the initial instance query is decomposed into frame-level box queries at the first decoder layer. The box queries are kept on each frame and serve as anchors without interacting with each other. The features extracted by box queries from each frame are aggregated to the instance query after each decoder layer, which is used for predicting dynamic mask head parameters. Then the mask head convolves the encoded feature maps to generate the mask sequences.  }
\label{fig:architecture}
\end{figure*}

\section{Method}

\subsection{Architecture}
The network architecture is visualized in Fig.~\ref{fig:architecture}. SeqFormer has a CNN backbone and a transformer encoder for extracting feature maps from each frame independently. 
Next, a transformer decoder is adapted to locate the instance sequences and generate a video-level instance representation.
Finally, three output heads are used for instance classification, instance sequences segmentation, and bounding box prediction, respectively.
 
\noindent\textbf{Backbone.}
Given an input video $x_v\in\mathbb{R}^{T\times3\times H \times W}$ with 3 color channels and $T$ frames of resolution $ H\times W$, the CNN backbone (\emph{e.g.}, ResNet~\cite{resnet}) extracts feature maps for each frame independently.

\noindent\textbf{Transformer Encoder.}
First, a $ 1\times 1 $ convolution is used to reduce the channel dimension of the all the feature maps to $C=256$, creating new feature maps $ \{\textbf{f}^{\prime}_t\}^T_{t=1}, {\textbf{f}}^{\prime}_t \in\mathbb{R}^{C\times H^{\prime} \times W^{\prime}}, t\in[1,T]$. After adding fixed positional encodings~\cite{detr}, the transformer encoder performs deformable attention~\cite{deformableDETR} on the feature maps, resulting in the output feature maps $ \{\textbf{f}_t\}^T_{t=1},$ with the same resolutions as the input. 
To perform attention mechanisms on each frame independently, we keep the spatial and temporal dimensions of feature maps rather than flattening them into one dimension.

\noindent\textbf{Query Decompose Transformer Decoder.}
Given a video, humans can effortlessly identify every instance and associate them through the video, despite the various appearance and changing positions on different frames. If an instance is hard to recognize due to occlusion or motion blur in some frames, humans can still re-identify it through the context information from other frames. In other words, for the same instance on different frames, humans treat them as a whole instead of individuals. This is the crucial difference between video and image instance segmentation. Motivated by this, we propose Query Decompose Transformer Decoder, which aims to learn a  more and robust video-level instance representation across frames.

We introduce a fixed number of learnable embeddings to query the features of the same instance from each frame, termed Instance Queries. Different from the instance queries corresponding to frame-level instances in VisTR~\cite{VisTR}, 
which has a fixed number of input queries hardcoded by video length and maximum number of instances, our instance queries correspond to video-level instances. Since the changing appearance and position of the instance, the model should attend to different exact spatial locations of each frame. 
To achieve this goal, we propose to decompose the instance query into $T$ frame-specific box queries, each of which serves as an anchor for retrieving and locating features on the corresponding frame.

At the first decoder layer, an instance query $\textbf{I}_q \in \mathbb{R}^{C}$ is used to query the instance features on features maps of each frame independently:
\begin{equation} \label{eq:1}
\textbf{B}^1_t = \text{DeformAttn}(\textbf{I}_q, \textbf{f}_t ),
\end{equation}
where $\textbf{B}^1_t\in \mathbb{R}^{C}$ is the box query on frame $t$ from the $1$-st decoder layer, and DeformAttn indicates deformable attention module in ~\cite{deformableDETR} . Given a query element and the frame feature map $\textbf{f}_t$, deformable attention only attends to a small set of key sampling points.
At the $l$-th $(l>1)$ layer, the box query $\textbf{B}^{l-1}_t$ from last layer is used as input:
\begin{equation} \label{eq:2}
\textbf{B}^l_t = \text{DeformAttn}(\textbf{B}^{l-1}_t, \textbf{f}_t ),
\end{equation}
and the instance query aggregates the temporal features by a weighted sum of all the box queries at the end of every decoder layers, where the weights are end to end learned upon the box embedding:
\begin{equation} \label{eq:sum}
\textbf{I}^l_q =   \frac{\sum^T_{t=1}\textbf{B}^l_t \times \text{FC}(\textbf{B}^l_t) }{\sum^T_{t=1}\text{FC}(\textbf{B}^l_t)}+\textbf{I}^{l-1}_q.
\end{equation}
After $N_d$ decoder layers, we get an instance query and $T$ box queries for each instance. The instance query is a shared video-level instance representation, and the box query contains the position information for predicting the bound box on each frame. 
We define the instance query $I_q^{N_d}$ and box queries $\{B_t^{N_d}\}^T_{t=1}$ from the last layer of decoder as output instance embedding and box embeddings  $ \{\textbf{BE}_t\}^T_{t=1}, \textbf{BE}_t\in \mathbb{R}^{N\times d} $.


\noindent\textbf{Output Heads.}
As shown in Fig.~\ref{fig:architecture}, we add mask head, box head, class head on the top of the decoder outputs.
A linear projection acts as the class head to produce the classification results. Given the instance embedding from the transformer decoder with index $\sigma(i)$, class head output a class probability of class $\textbf{c}_i$ (which may be $\varnothing$) as $\hat{p}_{\sigma(i)}(\textbf{c}_i)$ .

The box head is a 3-layer feed forward network (FFN) with ReLU activation function and a linear projection layer. For $\textbf{BE}_t$ of each frame, the FFN predicts the normalized center coordinates, height and width of the box w.r.t. the frame. Thus, for the instance with index $\sigma(i)$, we denote the predicted box sequence as 
$\hat{\textbf{b}}_{\sigma(i)} = \{\hat{\textbf{b}}_{(\sigma(i),1)}, \hat{\textbf{b}}_{(\sigma(i),2)}, ..., \hat{\textbf{b}}_{(\sigma(i),T)} \}$.

As for mask head, we leverage dynamic convolution~\cite{tian2020conditional} as our mask head. 
The output instance embedding of decoder contains the information of instance on all frames, thus it can be regarded as a more robust instance representation.
We can use instance embedding to efficiently generate the entire mask sequences.
First, a 3-layer FFN encodes the instance embedding into parameters $\omega_i$ of mask head with index $\sigma(i)$, which has three $1 \times 1$ convolution layers. 
The instances with the same identity on different frames share the same mask head parameters, which makes the segmentation very efficient.
Each convolution layer has 8 channels and uses ReLU as the activation function except for the last one, following ~\cite{tian2020conditional}. 
As shown in Fig.~\ref{fig:architecture}, there is a mask branch that provides the feature maps for mask head to predict instance masks. We employ an FPN-like architecture as the mask branch to make use of multi-scale feature maps from transformer encoder and generate feature maps sequences $\{\hat{\textbf{F}}^1_{\text{mask}}, \hat{\textbf{F}}^2_{\text{mask}},...,\hat{\textbf{F}}^T_{\text{mask}}\}$ that are $\frac{1}{8}$ of the input resolution and have 8 channels for each frame independently. Then the feature map $\hat{\textbf{F}}^t_{\text{mask}}$ is concatenated with a map of the relative coordinates from center of $\hat{\textbf{b}}_{(\sigma(i),t)}$ in corresponding frames to provide a location cue for predicting the instance mask.
Thus we get the $ \{\textbf{F}^t_{\text{mask}}\}^T_{t=1},\textbf{F}^t_{\text{mask}}\in\mathbb{R}^{10\times \frac{H}{8} \times \frac{W}{8}}$. 
The mask head performs convolution on these high-resolution sequence feature maps $\textbf{F}^t_{\text{mask}}$ to predict the mask sequences:
\begin{equation} \label{eq:dynamic_conv}
\{\textbf{m}^t_i\}^T_{t=1} = \{\text{MaskHead}(\textbf{F}^t_{\text{mask}},\omega_i) \}^T_{t=1},
\end{equation}
where MaskHead performs three-layer $1\times1$ convolution on given feature maps with the kernels reshaped from $\omega$.
By sharing the same mask head parameters for instances with the same identity on different frames, our method can efficiently perform instance segmentation on each frame.
Similar to DETR~\cite{detr}, we add output heads and Hungarian loss after each decoder layer as an auxiliary loss to supervise the training stage.

\subsection{Instance Sequences Matching and Loss}

Our method predicts a fixed-size set of $N$ predictions in a single pass through the decoder, and $N$ is set to be significantly larger than the number of instances in a video. To train our network, we first need to find a bipartite graph matching between the prediction and the ground truth. Let $\textbf{y}$ denotes the ground truth set of video-level instance, and $\hat{\textbf{y}}_i = \{ \hat{\textbf{y}}_i\}^N_{i=1}$ denotes the predicted instance set. Each element $i$ of the ground truth set can be seen as $\textbf{y}_i = \{ \textbf{c}_i,(\textbf{b}_{i,1}, \textbf{b}_{i,2},...,\textbf{b}_{i,T}) \}$,
where $\textbf{c}_i$ is the target class label including $\varnothing$, and $\textbf{b}_{i,t} \in [0,1]^4$ is a vector that defines ground truth bounding box center coordinates and its relative height and width in the frame $t$. For the predictions of instance with index $\sigma(i)$, we take the output of class head  $\hat{p}_{\sigma(i)}(\textbf{c}_i)$ and predicted bounding box $\hat{\textbf{b}}_{\sigma(i)}$. Then we define the pair-wise matching cost between ground truth $y_i$ and a prediction with index $\sigma(i)$.
\begin{equation} \label{eq:5}
\mathcal{L}_{\text{match}}(\textbf{y}_i,\hat{\textbf{y}}_{\sigma(i)}) = -\hat{p}_{\sigma(i)}(\textbf{c}_i) + \mathcal{L}_{\text{box}}(\textbf{b}_i,\hat{\textbf{b}}_{\sigma(i)}),
\end{equation}
where $\textbf{c}_i \neq \varnothing$. 
Note that Eq.~\ref{eq:5} does not consider the similarity between mask prediction and mask ground truth, as such mask-level comparison is computationally expensive.
To find the best assignment of a ground truth to a prediction, we search for a permutation of N elements $\sigma\in S_n$ with the lowest cost:
\begin{equation} \label{eq:6}
\hat{\sigma} = \mathop{\arg\min}\limits_{\sigma \in S_n} \sum^N_i \mathcal{L}_{\text{match}}(\textbf{y}_i,\hat{\textbf{y}}_{\sigma(i)}).
\end{equation}

Following prior work\cite{detr,VisTR}, the optimal assignment is computed with the Hungarian algorithm~\cite{kuhn1955hungarian}. Given the optimal assignment  $\hat{\sigma}$ , we use Hungarian loss for all matched pairs to train our network:
\begin{equation} \label{eq:Loss_function}
\begin{aligned}
\mathcal{L}_{\text{Hung}}(\textbf{y},\hat{\textbf{y}})=\sum^N_{i=1} &\left[-\log\hat{p}_{\hat{\sigma}(i)}(\textbf{c}_i) + \mathds{1}_{\{\textbf{c}_i \neq \varnothing\}} \mathcal{L}_{\text{box}}(\textbf{b}_i,\hat{\textbf{b}}_{\hat{\sigma}}(i)) \right. \\
&\left. +\mathds{1}_{\{\textbf{c}_i \neq \varnothing\}}  \mathcal{L}_{\text{mask}}(\textbf{m}_i,\hat{\textbf{m}}_{\hat{\sigma}}(i)) \right].
\end{aligned}
\end{equation}
For $\mathcal{L}_{\text{box}}$, we use a linear combination of the  $\mathcal{L}_1$ loss and the generalized IoU loss~\cite{rezatofighi2019generalized}. The mask sequences $\{m^t_i\}^T_{t=1} $ from mask head with $\frac{1}{8}$ of the video resolution which may loss some details, thus we upsample the predicted mask to $\frac{1}{4}$ of the video resolution, and downsample the ground truth mask to the same resolution for mask loss, following~\cite{tian2020conditional}. The mask loss $\mathcal{L}_{\text{mask}}$ is defined as a combination of the Dice~\cite{milletari2016v} and Focal loss~\cite{lin2017focal}. 
We calculate box loss and mask loss on each frame and take the average for Hungarian loss.

\section{Experiment}
\subsection{ Datasets and Metrics }
We evaluate our method on YouTube-VIS 2019~\cite{yang2019video} and YouTube-VIS 2021~\cite{ytvis21dataset} datasets. YouTube-VIS 2019 is the first and largest dataset for video instance segmentation, which contains 2238 training, 302 validation, and 343 test high-resolution YouTube video clips. It has a 40-category label set and 131k high-quality instance masks. In each video, objects with bounding boxes and masks are labeled every five frames. YouTube-VIS 2021 is an improved and extended version of YouTube-VIS 2019 dataset, it contains 3,859 high-resolution videos and 232k instance annotations. The newly added videos in the dataset include more instances and frames.

Video instance segmentation is evaluated by the metrics of average precision (AP) and average recall (AR). Different from image instance segmentation, each instance in a video contains a sequence of masks. 
To evaluate the spatio-temporal consistency of the predicted mask sequences, the IoU computation is carried out in the spatial-temporal domain. This requires a model not only to obtain accurate segmentation and classification results at frame-level but also to track instance masks between frames accurately.

\subsection{ Implementation Details}

\noindent\textbf{Model settings.}
ResNet-50~\cite{resnet} is used as our backbone network unless otherwise specified. Similar to ~\cite{deformableDETR}, we use the features from the last three stages as \{C3, C4, C5\} in ResNet, which correspond to the feature maps with strides \{8, 16, 32\}. And adding the lowest resolution feature map C6 obtained via a 3 × 3 stride 2 convolution on the C5.
We set sampled key numbers K=4 and eight attention heads for deformable attention modules.
We use six encoder and six decoder layers of hidden dimension 256 for the transformer, and the number of instance queries is set to 300.

\noindent\textbf{Training.}
We used AdamW~\cite{AdamW} optimizer with base learning rate of $2\times 10^{-4}$, $\beta1=0.9$ , $\beta2=0.999$, and weight decay of $10^{-4}$. Learning rates of the backbone and linear projections used for deformable attention modules are multiplied by a factor of 0.1.
We first pre-train the model on COCO~\cite{coco} by setting the number of input frames  $T=1$.
Given the pretrained weights, we train our models on the YouTube-VIS dataset with input frames  $T=5$ sampled from the same video. 

The training data of the YouTube-VIS dataset is not sufficient, which makes a model prone to overfitting. To address this problem, we adopt 80K training images in the COCO for compensation, following~\cite{athar2020stem,ProposeReduce}. 
We only use the images with 20 overlapping categories in COCO and augment them with  $\pm 10^\circ $ rotation to generate a five-frame pseudo video.
We train our model on the mixed dataset including COCO and the video dataset for 12 epochs, and the learning rate is decayed at the 6-th and 10-th epoch by a factor of 0.1. The input frame sizes are downsampled so that the longest side is at most 768 pixels.
The model is implemented with PyTorch-1.7 and is trained on 8 V100 GPUs of 32G RAM, with 2 video clips per GPU. 

\noindent\textbf{Inference. }
SeqFormer is able to model a video of arbitrary length without grouping frames into subsequences. 
We take the whole video as input during inference, which is downscaled to 360p, following MaskTrack R-CNN~\cite{yang2019video}.
SeqFormer learns a video-level instance representation used for dynamic segmentation on each frame and classification, and the box sequences are generated by the decoder. Thus, no post-processing is needed for associating instances.

\begin{table}[t]
\begin{center}
\caption{Quantitative results of video instance segmentation on YouTube-VIS 2019 validation set.
The result with superscript ``\dag" is obtained without coco joint training.
The best results with the same backbone are in \textbf{bold}.}
\label{table:mainytvis}
\begin{tabular}{l|l|c|c|ccccc}
\hline\noalign{\smallskip}
Backbone   &Method  &Params &FPS &$\rm AP$    &$\rm AP_{50}$  &$\rm AP_{75}$ &$\rm AR_{1}$  &$\rm AR_{10}$  \\
\noalign{\smallskip}
\hline
\noalign{\smallskip}
\multirow{11}{*}{ResNet-50}
 &MaskTrack R-CNN~\cite{yang2019video}  &58.1M &20.0 &30.3 &51.1 &32.6 &31.0 &35.5  \\
 &STEm-Seg~\cite{athar2020stem}  &50.5M   &7.0 &30.6 &50.7 &33.5 &37.6 &37.1 \\
 &SipMask~\cite{cao2020sipmask}  &33.2M  &30.0 &33.7 &54.1 &35.8 &35.4 &40.1   \\
 &CompFeat~\cite{fu2020compfeat} &-  &- &35.3 &56.0 &38.6 &33.1 &40.3   \\
 &SG-Net~\cite{liu2021sg} &-  &- &34.8 &56.1 &36.8 &35.8 &40.8  \\
&VisTR~\cite{VisTR}   &57.2M  &69.9 &36.2 &59.8 &36.9 &37.2 &42.4    \\
 &MaskProp~\cite{MaskProp}  &-   &- &40.0 &\ \ \ -    &42.9 &\ \ \   -   &\ \ \  -    \\  
 &CrossVIS~\cite{yang2021crossover} &37.5M &39.8 &36.3 &56.8 &38.9 &35.6 &40.7   \\
 &Propose-Reduce~\cite{ProposeReduce}  &69.0M  &- &40.4 &63.0 &43.8 &41.1 &49.7   \\
 &IFC~\cite{IFC} &39.3M &107.1 &42.8 &65.8 &46.8 &43.8 &51.2  \\  
 &\textbf{SeqFormer}$^{\dag}$   &49.3M   &72.3  &45.1 &66.9 &50.5 &\textbf{45.6} &54.6 \\
 &\textbf{SeqFormer}   &49.3M   &72.3  &\textbf{47.4} &\textbf{69.8} &\textbf{51.8} &{45.5} &\textbf{54.8}  \\  
\hline
\multirow{9}{*}{ResNet-101}
 &MaskTrack R-CNN~\cite{yang2019video}  &77.2M   &-  &31.8 &53.0 &33.6 &33.2 &37.6  \\
 &STEm-Seg~\cite{athar2020stem} &69.6M &- &34.6 &55.8 &37.9 &34.4 &41.6 \\
 &SG-Net~\cite{liu2021sg}  &- &-  &36.3 &57.1 &39.6 &35.9 &43.0  \\
 &VisTR~\cite{VisTR}  &76.3M   &57.7  &40.1 &64.0 &45.0 &38.3 &44.9  \\
 &MaskProp~\cite{MaskProp} &- &-  &42.5 &\ \ \  -   &45.6 &\ \ \  -   &\ \ \  -   \\  
 &CrossVIS~\cite{yang2021crossover}  &56.6   &35.6  &36.6 &57.3 &39.7 &36.0 &42.0 \\
 &Propose-Reduce~\cite{ProposeReduce}  &88.1M   &-  &43.8 &65.5 &47.4 &43.0 &53.2  \\
 &IFC~\cite{IFC}  &58.3M  &89.4  &44.6 &69.2 &49.5 &44.0 &52.1  \\  
 &\textbf{SeqFormer}   &68.4M   &64.6  &\textbf{49.0} &\textbf{71.1} &\textbf{55.7}   &\textbf{46.8}  &\textbf{56.9}  \\  
\hline
\multirow{3}{*}{ResNeXt-101}
 &MaskProp~\cite{MaskProp} &-  &- &44.3 &\ \ \  -   &48.3 &\ \ \  - &\ \ \  -  \\  
 &Propose-Reduce~\cite{ProposeReduce} &127.1M  &- &47.6 &71.6 &51.8 &46.3 &56.0   \\ 
  &\textbf{SeqFormer} &112.7M  &30.7 &\textbf{51.2}  &\textbf{75.3}  &\textbf{58.0} &\textbf{46.5} &\textbf{57.3}  \\  
\hline
 Swin-L   &\textbf{SeqFormer}  &220.0M   &27.7  &\textbf{59.3}  &\textbf{82.1}  &\textbf{66.4}  &\textbf{51.7}  &\textbf{64.4}  \\ 
\hline

\end{tabular}
\end{center}
\end{table}

\subsection{ Main Results }

\begin{figure*}[tb]
\centering
\includegraphics[width=0.9 \linewidth]{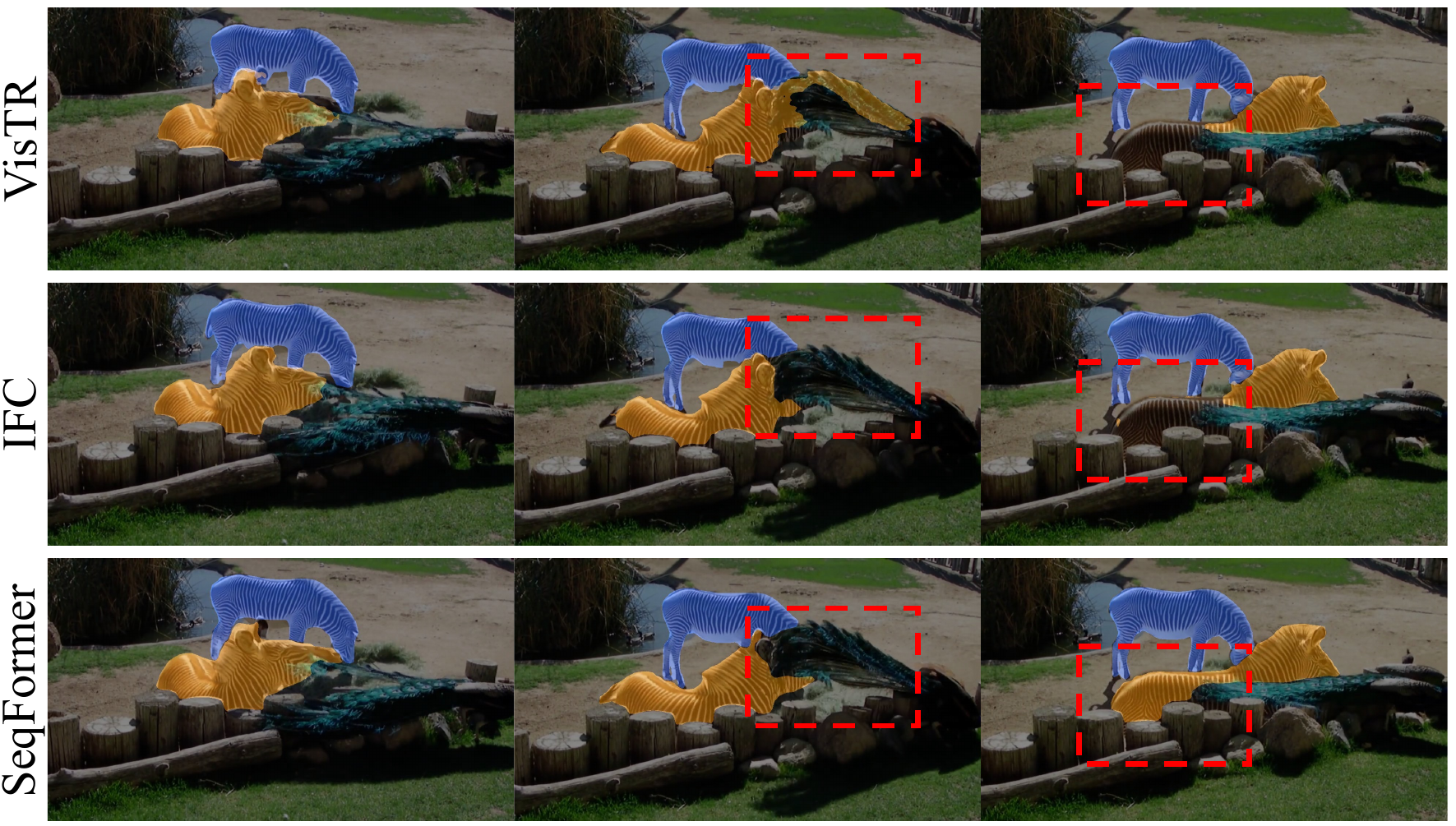}
\caption{Qualitative comparisons with other methods.}
\label{fig:visualization_compare}
\end{figure*}

The comparison of SeqFormer with previous state-of-the-art methods on YouTube-VIS 2019 are listed in Table~\ref{table:mainytvis}. 
MaskProp~\cite{MaskProp} and ProposeReduce~\cite{ProposeReduce} are the state-of-the-art methods, which take a strong backbone to extract spatial features and use mask propagation to improve the segmentation and tracking, but suffer from low inference speed.
We list the methods with different backbones for fair comparison. It can be observed that SeqFormer significantly surpasses all the previous best reported results by at least 4 AP with the same backbone. 
Training our model with coco pseudo videos improves the AP from 45.1 to 47.4. 
SeqFormer with a ResNet-50 backbone can even achieve competitive performance against state-of-the-art methods with a ResNeXt-101 backbone. By adopting Swin transformer~\cite{swin} as our backbone without further modifications, SeqFormer can first achieve 59.3 AP on this benchmark, outperforming the best previous results by a large margin of 11.7 AP. 
To understand the runtime efficiency, we measure FPS of SeqFormer excluding the data loading process of multiple images on NVIDIA Tesla V100. With an input size of 360p and a ResNet-50 backbone on YouTube-VIS, the inference FPS is 72.3. While surpassing the stateof-the-art AP by a large margin, SeqFormer is the second fast one following IFC.
An example of qualitative comparison with previous methods is given in Fig.~\ref{fig:visualization_compare}, the mask predictions of SeqFormer are more stable over time.
Please refer to the Sup. Mat. for more qualitative results.
We also evaluate our approach on the recently introduced YouTube-VIS 2021 dataset, which is a more challenging dataset with more videos and a higher number of instances and frames. As shown in Table~\ref{table:main_ytvis21}, SeqFormer achieves 40.5 AP with a ResNet-50 backbone, surpassing previous methods by 3.9 AP.
We believe that our effective method will serve as a strong baseline on these benchmarks and facilitate future research in video instance segmentation.


\setlength{\tabcolsep}{4pt}
\begin{table}[tb]
\begin{center}
\caption{Quantitative results of video instance segmentation on YouTube-VIS 2021 validation set. The best results with the same backbone are in \textbf{bold}.}
\label{table:main_ytvis21}
\begin{tabular}{llccccccc}
\hline\noalign{\smallskip}
Backbone   &Method  &$\rm AP$    &$\rm AP_{50}$  &$\rm AP_{75}$ &$\rm AR_{1}$  &$\rm AR_{10}$  \\
\noalign{\smallskip}
\hline
\noalign{\smallskip}
\multirow{5}{*}{ResNet-50}
&MaskTrack R-CNN~\cite{yang2019video}     &28.6 &48.9 &29.6 &26.5 &33.8 \\
&SipMask~\cite{cao2020sipmask}        &31.7 &52.5 &34.0 &30.8 &37.8    \\
&CrossVIS~\cite{yang2021crossover}        &34.2 &54.4 &37.9 &30.4 &38.2  \\
&IFC~\cite{IFC}            &36.6 &57.9 &39.3 &- &-  \\
&\textbf{SeqFormer}        &\textbf{40.5} &\textbf{62.4} &\textbf{43.7} &\textbf{36.1} &\textbf{48.1}  \\  
\hline
Swin-L &\textbf{SeqFormer}       &\textbf{51.8} &\textbf{74.6} &\textbf{58.2} &\textbf{42.8} &\textbf{58.1} \\
\hline
\end{tabular}
\end{center}
\end{table}

\subsection{ Ablation Study }
This section conducts extensive ablation experiments to study the effects of different settings in our proposed method. All the ablation experiments are conducted with the ResNet-50 backbone and training on YouTube-VIS 2019 dataset rather than the mixed dataset.

\setlength{\tabcolsep}{4pt}
\begin{table}[tb]
\begin{center}
\caption{Instance query decomposition. Decomposing instance query into frame-level box queries is critical for SeqFormer.}
\label{query_decompose}
\begin{tabular}{lccccc}
\hline\noalign{\smallskip}
Decompose    &$\rm AP$  &$\rm AP_{50}$  &$\rm AP_{75}$ &$\rm AR_{1}$  &$\rm AR_{10}$  \\
\noalign{\smallskip}
\hline
\noalign{\smallskip}
w/o   &34.1  &53.7  &34.9  &34.8  &40.9\\
w     &45.1 &66.9 &50.5 &45.6 &54.6 \\
\hline
\end{tabular}
\end{center}
\end{table}
\setlength{\tabcolsep}{1.4pt}

\begin{figure}[tb]
\centering
\includegraphics[width = 0.9\linewidth]{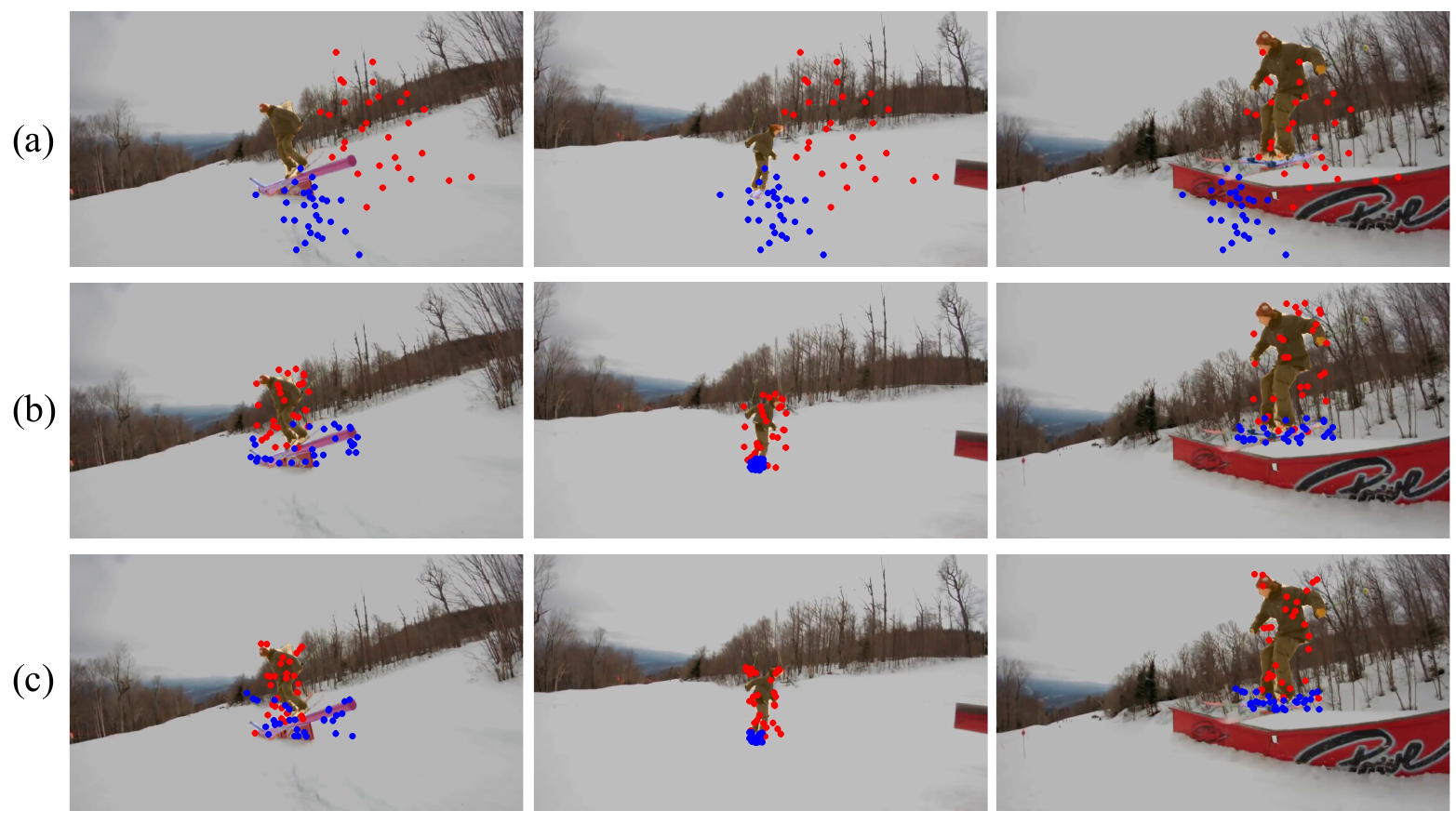}
\caption{The sampling points from the first decoder layer is shown in (a), which is coarse and inaccurate. The refined accurate sampling points from the second and last decoder layer are shown in (b) and (c). }
\label{coase_to_fine}
\end{figure}

\noindent\textbf{Instance query decomposition.}
Instance query decomposition plays an important role in our method. Since an instance may have different positions on each frame, the iterative refinement of the spatial sampling region should be performed independently on each frame. 
To keep the temporal consistency of instances, we use the temporal-shared instance query for deformable attention and get box queries for each frame. The box queries will be kept through all the decoder layers and serve as frame anchors for the same instance. Experiments of models without box queries and using the shared instance query for each decoder layer are presented in Table~\ref{query_decompose}. 
The model without query decomposition manages to achieve only 34.1 AP. 
It is because the query controls the sampling region of deformable attention. Using the same instance query for each frame will result in the same spatial sampling region on each frame, as shown in Fig.~\ref{coase_to_fine} (a), which is inaccurate and insufficient for video-level instance representation. We further visualize the sampling points of the second and the last decoder layers in Fig.~\ref{coase_to_fine} (b) and (c). The box queries decoupled from instance query serve as anchors for locating features and iteratively refining the sampling region on the current frame. It can be seen that SeqFormer attends to locations following with the movement of instance through the video in a coarse-to-fine manner. Please refer to the Sup. Mat. for more visualization of sampling points.

\noindent\textbf{Spatial temporal dimensions.}
Previous transformer-based methods~\cite{VisTR,IFC} flatten the spatial and temporal dimensions of video features into one dimension for the transformer decoder. We argue that the temporal dimension should not be flattened with spatial dimensions, since it was recognized that the 2D space domain and 1D time domain have different characteristics and should be intuitively handled in a different way~\cite{zhao2018trajectory}. Thus, we retain the complete 3D spatio-temporal dimensions and perform explicit region sampling and information aggregation on all frames. In this experiment, we study the effect of this architecture by replacing deformable transformer with vanilla transformer and flattening the spatial and temporal dimensions, termed as `flatten' in Table~\ref{ablation_decoder}. For fair comparison, we use single-scale deformable attention as the baseline, termed as `single-scale', which use the same scale feature map with `flatten', the default setting termed as `multi-scale'.
By keeping spatial-temporal dimensions of video features, the AP increased from 35.1 to 42.5. The use of multi-scale feature maps can only improve 2.6 AP, which proves that the success of our method mainly comes from the preservation of the temporal dimension and the explicit spatial sampling.

\setlength{\tabcolsep}{4pt}
\begin{table}[tb]
\begin{center}
\caption{Spatial and temporal dimensions. Keeping spatial-temporal feature dimensions and performing instance feature capture independently on different frames brings about 7.4 AP gains. Multi-scale feature maps can further bring 2.6 AP. }
\label{ablation_decoder}
\begin{tabular}{lccccc}
\hline\noalign{\smallskip}
Feature    &$\rm AP$  &$\rm AP_{50}$  &$\rm AP_{75}$ &$\rm AR_{1}$  &$\rm AR_{10}$  \\
\noalign{\smallskip}
\hline
\noalign{\smallskip}
flatten          &35.1  &56.8  &35.6  &38.1  &41.8\\
single-scale     &42.5  &64.6  &46.5  &41.5  &50.9\\
multi-scale      &45.1  &66.9  &50.5  &45.6  &54.6\\
\hline
\end{tabular}
\end{center}
\end{table}
\setlength{\tabcolsep}{1.4pt}

\noindent\textbf{Aggregation of temporal information.}
The frame-level box queries and the predicted boxes can align the instance features from all frames, there are several ways to aggregate the aligned features into the instance query.
We conduct an experiment to evaluate the different aggregation ways for these features, as shown in Table~\ref{ablation_combinations}.
In the ‘sum' setting, the features from different frames are directly added together as the instance feature of this decoder layer. In the ‘average' setting, the feature on each frame is averaged as the instance feature. In the ‘weighted-sum' setting, we apply a softmax layer and a fully-connected layer on box embeddings to get the weights of each frame, and the features are aggregated in a weighted sum in Eq.~\ref{eq:sum}. 
The result is 30.6 AP and 43.2 AP for `sum' and `average' settings respectively. Direct summation will cause the value to be unstable with different frame numbers. Since some instances only appear in a few frames, directly averaging features from all frames may cause the information to be diluted. Please refer to the Sup. Mat. for more details and visualization of different frame weights.

\setlength{\tabcolsep}{4pt}
\begin{table}[tb]
\begin{center}
\caption{Temporal information aggregation. Weighted sum brings a performance gain of 1.9 in AP.}
\label{ablation_combinations}
\begin{tabular}{lccccc}
\hline\noalign{\smallskip}
Aggregation    &$\rm AP$    &$\rm AP_{50}$  &$\rm AP_{75}$ &$\rm AR_{1}$  &$\rm AR_{10}$  \\
\noalign{\smallskip}
\hline
\noalign{\smallskip}
sum   &30.6  &44.5  &34.3  &37.2  &45.0 \\
average   &43.2  &65.2  &48.5  &43.4  &52.8 \\
weighted-sum   &45.1 &66.9 &50.5 &45.6 &54.6\\
\hline
\end{tabular}
\end{center}
\end{table}
\setlength{\tabcolsep}{1.4pt}

\noindent\textbf{Robust instance representation.}
Our decoder explicitly aligns and aggregates the information from each frame to learn a video-level instance representation.
In this experiment, we try to generate instance representation with fewer frames. We use the instance representation to generate a mask head and apply the mask head on each frame to get the mask sequences, as shown in Table~\ref{reduce_frames}. Surprisingly, with only one frame as input, the generated mask head can produce a competitive result of 38.1 AP. With five frames as input, the performance is only 0.5 AP worse than taking all frames as input. This result shows that the mask head learned by our method can generalize well to unseen frames.

\setlength{\tabcolsep}{4pt}
\begin{table}[tb]
\begin{center}
\caption{Fewer frames for instance representation.We evenly sample fewer frames from a video to generate the mask head.}
\label{reduce_frames}
\begin{tabular}{lccccc}
\hline\noalign{\smallskip}
Frames    &$\rm AP$    &$\rm AP_{50}$  &$\rm AP_{75}$ &$\rm AR_{1}$  &$\rm AR_{10}$  \\
\noalign{\smallskip}
\hline
\noalign{\smallskip}
1  &38.1  &58.3  &41.3  &38.7  &47.5   \\
3   &43.4  &65.4  &47.6  &42.4  &51.3  \\
5   &44.6  &66.5  &49.7  &44.8  &54.6  \\
10   &44.7  &66.9  &49.5  &44.3  &53.5 \\
all  &45.1  &66.9  &50.5 &45.6 &54.6   \\
\hline
\end{tabular}
\end{center}
\end{table}
\setlength{\tabcolsep}{1.4pt}

\section{Conclusion}
In this paper, we have proposed an effective transformer architecture for video instance segmentation, named SeqFormer, which performs attention mechanisms on each frame independently and learns a shared powerful instance query for each video-level instance. With the proposed instance query decomposition, our network can align the instance features and naturally tackle the instance tracking without additional tracking branches or post-processing.
We demonstrated that our method surpasses all state-of-the-art methods by a large margin.  We believe that our  neat and efficient approach will serve as a strong baseline for future research in video instance segmentation.

~\\
\noindent\textbf{Acknowledgments.}
We thank Xiaoding Yuan for the support and discussions about implementation details.
We thank the anonymous reviewers for their efforts and valuable feedback to improve our work.

%
%
\bibliographystyle{splncs04}
\bibliography{egbib}

\clearpage
\input{appendix.tex}

\end{document}

%% file: appendix.tex
\appendix
\section{Appendix}

\begin{figure*}[h]
\centering
\includegraphics[width=1 \linewidth]{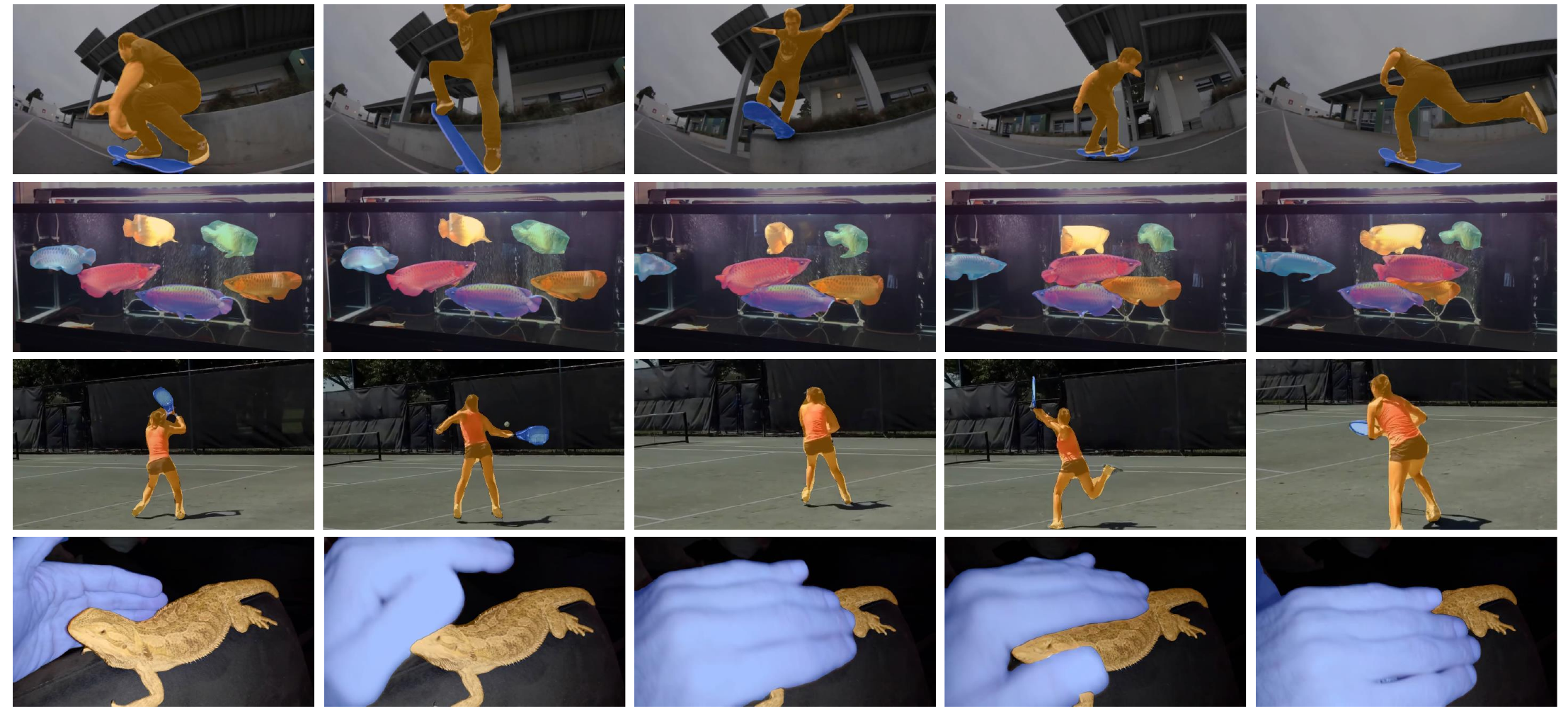}
\caption{Visualization of SeqFormer on the YouTube-VIS 2019 validation dataset. The first row shows the instances with various poses. The second row shows the case of a lot of similar instances that are close together with overlapping.
The third row shows the situation where an instance reappears after being occluded while in motion. The last row shows an instance severely occluded by the other instance. The same colors depict the mask sequences of the same instances}
\label{fig:visualization}
\end{figure*}

\begin{figure}[h]
\centering
\includegraphics[width = 1\textwidth]{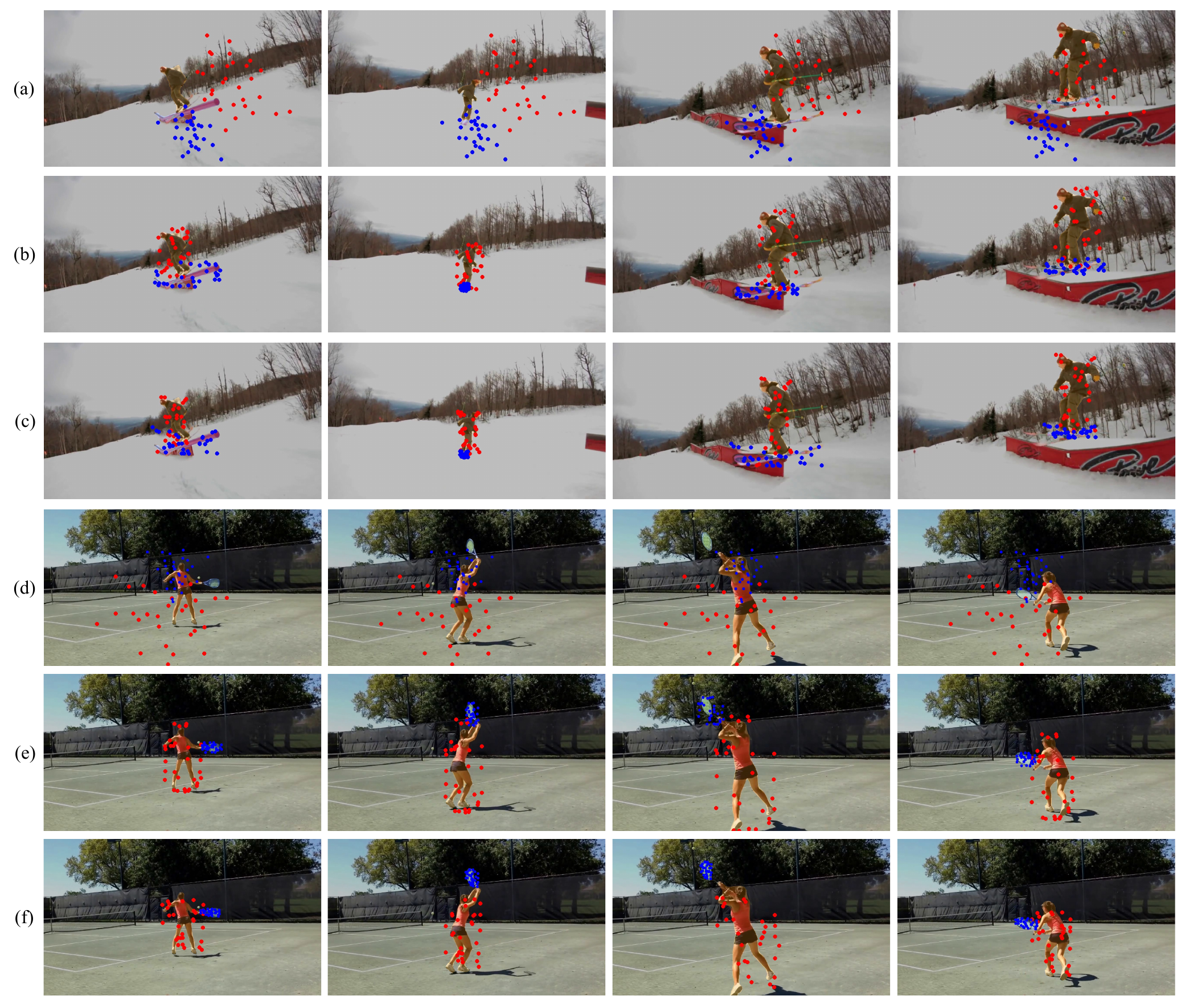}
\caption{Visualization of attention. We draw the sampling points that the deformable attention attends to. The four frames in each row are from the same video. Each sampling point is marked as a filled circle whose color indicates its corresponding instance identity. (a) and (d) show the sampling points from the first decoder layer. (b) and (e) show the sampling points from the second decoder layer. (c) and (f) show the sampling points from the last decoder layer.}
\label{fig:sampling_points}
\end{figure}

\subsection{Visualization}
\label{sec:visualization}

In Fig.~\ref{fig:visualization}, we visualize the results of SeqFormer with four challenging cases. It can be seen that SeqFormer can handle these situations well.
In Fig.~\ref{fig:sampling_points}, we show more qualitative results of the intermediate attention of transformer decoder. 
Since the same initial instance query is used to predict sampling points for each frame in the first decoder layer (Eq.1), the distribution of sampling points on each frame is the same in Fig.~\ref{fig:sampling_points} (a) and (d). 
After that, the initial instance query is decomposed into frame-level box queries that are kept and maintained independently on each frame.
Starting from the second layer of the SeqFormer decoder, the box query is used to predict the sampling points of the current frame, and the sampled features are used to refine the box query for the next decoder layer.
By doing so, SeqFormer attends to different spatial locations following the motion of the instance in a coarse-to-fine manner.

\begin{figure}[h]
\centering
\includegraphics[width = 1\textwidth]{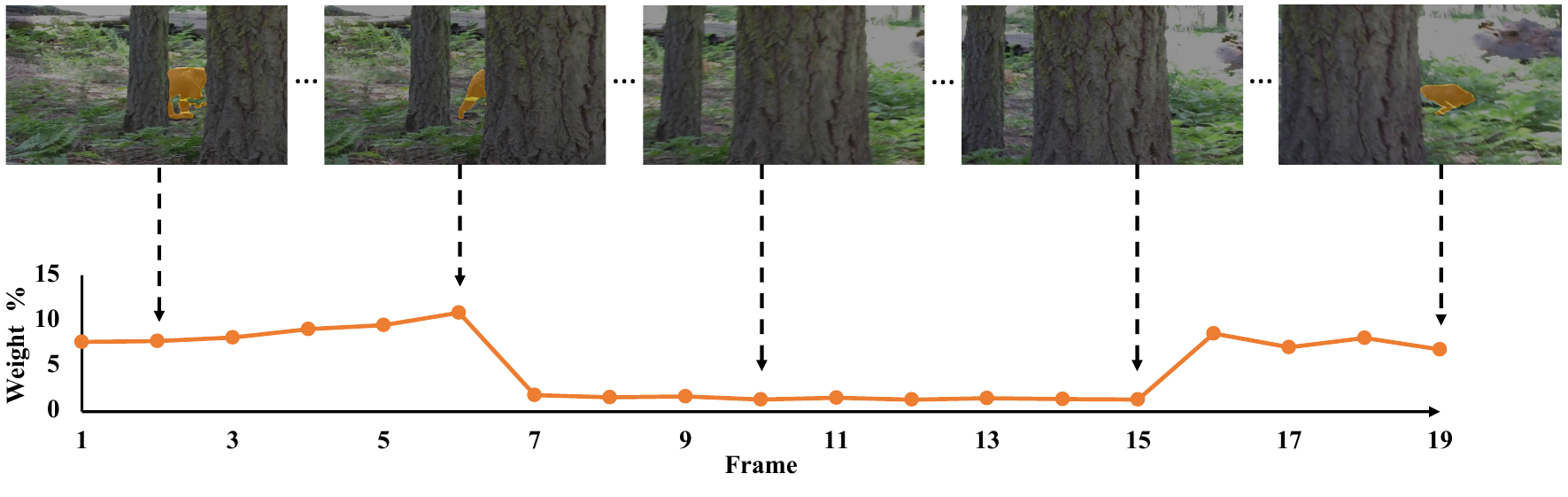}
\caption{Visualization of the normalized softmax weights and the corresponding frames.}
\label{fig:temporal}
\end{figure}

\subsection{Aggregation of Temporal Information}
\label{sec:aggregation}

SeqFormer is able to attend to different spatial locations following the motion of the instance. 
The aligned features are aggregated into an instance query to generate a video-level instance representation.
However, an instance may not appear in every frame due to occlusion and camera motion.
The features from frames without instance are useless or even harmful.
To address this, SeqFormer aggregates temporal features in a weighted manner, where the weights are learned upon the box queries in Eq.3. 
We visualize the learned weights and the corresponding frames in Fig.~\ref{fig:temporal}.
It can be seen that the features from frames without instance have lower weights.

\subsection{Qualitative Comparisons}
\begin{figure}[t]
\centering
\includegraphics[width = 0.9\textwidth]{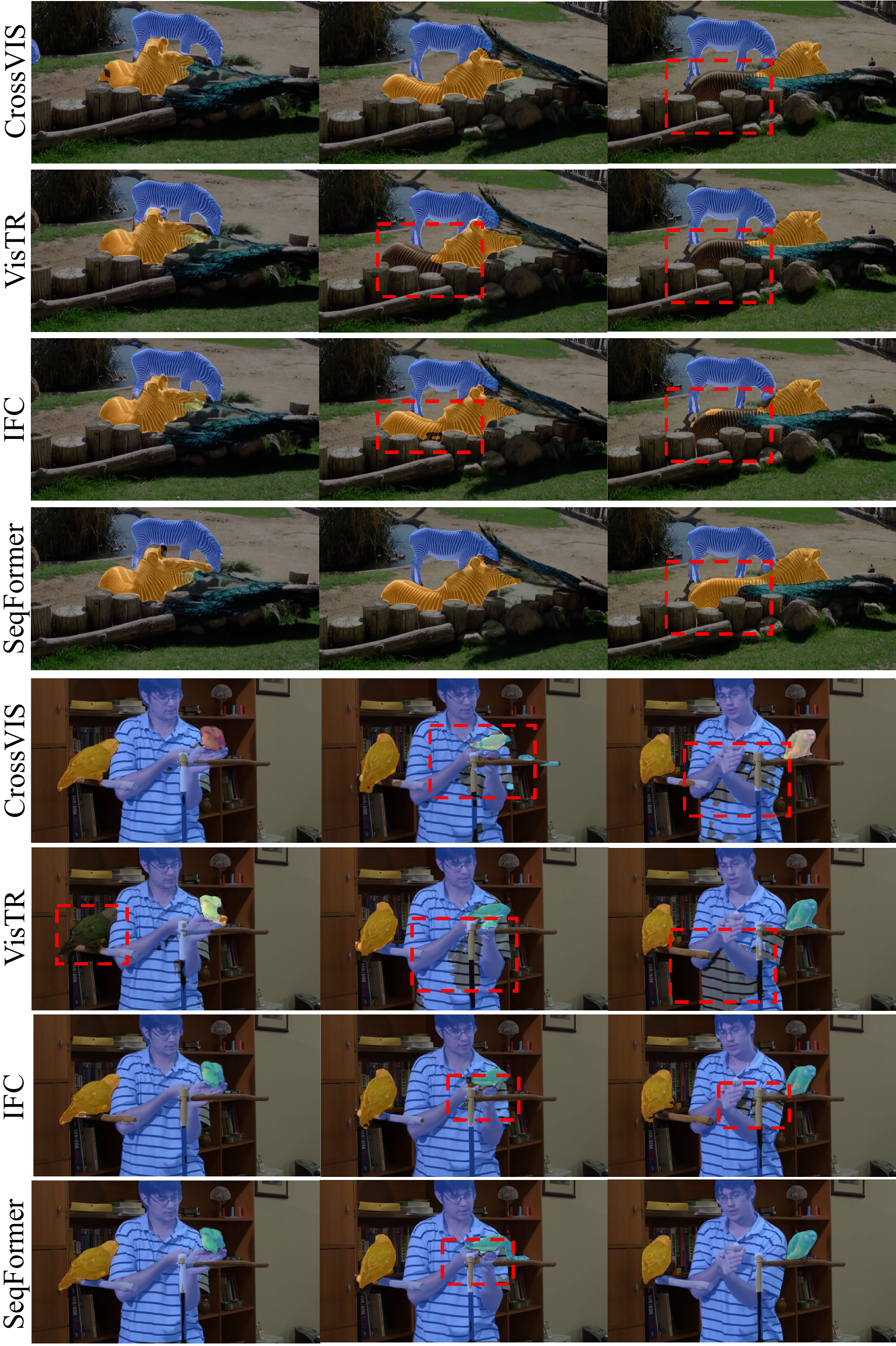}
\caption{Qualitative comparisons with other methods on YouTube-VIS 2019.
All methods use ResNet-50 backbone.
The three frames in each row are from the same video. 
The mask predictions of SeqFormer are more stable over time.
Best viewed in color.}
\label{fig:compare}
\end{figure}

We provide some qualitative comparisons with other methods in Fig.~\ref{fig:compare}, the mask predictions of SeqFormer are more stable over time. More video results and comparisons can be found in the rest of the supplementary material.

\subsection{Clip Matching}
Our model can be extended to per-clip model through clip matching algorithm to handle long videos.
Specifically, we divide long videos into clips with overlapping frames, and match clip-level instance masks by calculate the matching scores which are space-time soft IoU of overlapping frames, following IFC.
We evaluate our method by varying the length of clips in Table~\ref{clip_matching}. 
Our method still achieves competitive performance but slightly worse when evaluating in the clip-wise manner.
This manner makes our method handle very long videos with limited computational resources and have a wider range of application scenarios.

\setlength{\tabcolsep}{4pt}
\begin{table}[h]
\begin{center}
\caption{Evaluating SeqFormer in a clip-wise manner. }
\label{clip_matching}
\begin{tabular}{ccccc}
\hline\noalign{\smallskip}
Clip Length   &Whole &20 &15 &10    \\
\noalign{\smallskip}
\midrule
AP  &45.1 &44.8  &43.6  &42.0\\
\hline
\end{tabular}
\end{center}
\end{table}
\setlength{\tabcolsep}{1.4pt}